\documentclass[letterpaper, 10 pt, conference]{ieeeconf}  % Comment this line out if you need a4paper

\usepackage{comment}
\usepackage{bm}

\usepackage{color}
\usepackage{microtype} % Improved Tracking and Kerning

\usepackage{algpseudocode}

\RequirePackage{suffix}
\RequirePackage{soul}

\newcommand{\neweq}[2]{
\begin{equation}
 \label{e:#1}
 #2
\end{equation}
}

\WithSuffix\newcommand\neweq*[1]{
	$$
		#1
	$$
}

\newcommand{\Eq}[1]{Equation \ref{e:#1}}

\RequirePackage{float}
\restylefloat{table}

\RequirePackage{amsfonts}
\RequirePackage{amsmath}

\RequirePackage{graphicx}
\RequirePackage{wrapfig}
\RequirePackage{float}
\RequirePackage{subfig}
\newcommand{\leftfig}[3]{
\begin{figure}[h]
	\includegraphics[width=.5\textwidth]{#1}
	\caption{#3}
	\label{f:#2}
\end{figure}
}

\newcommand{\scaledfig}[4]{
	\begin{figure}[]
		{\centering
			\includegraphics[width=#4\textwidth]{#1}
			\caption{#3}
			\vspace{-20pt}
			\label{f:#2}
		}
	\end{figure}
}

\newcommand{\fig}[1]{Fig. \ref{f:#1}}

\newcommand{\Figure}[1]{Figure \ref{f:#1}}

\long\def\@makecaption#1#2{%
  \vskip\abovecaptionskip
  \sbox\@tempboxa{\footnotesize\bfseries #1\@. #2}%
  \ifdim \wd\@tempboxa >\hsize
    \footnotesize\bfseries #1\@. #2\par
  \else
    \global \@minipagefalse
    \hb@xt@\hsize{\hfil\box\@tempboxa\hfil}%
  \fi
  \vskip\belowcaptionskip}
\def\fps@figure{htbp}
\def\fps@table{htbp}

\newcommand*\colvec[1]{\begin{pmatrix}#1\end{pmatrix}}
\newcommand*\rowvec[1]{\begin{pmatrix}#1\end{pmatrix}^{T}}

\IEEEoverridecommandlockouts                              % This command is only needed if 
\title{Extrinisic Calibration of a Camera-Arm System Through Rotation Identification}

\author{Steve McGuire$^\ast$, Christoffer Heckman,  Daniel Szafir, Simon Julier, and Nisar Ahmed$^\ast$
\thanks{S.\ McGuire and N.\ Ahmed are with the Department of Aerospace Engineering Sciences, University of Colorado at Boulder. C.\ Heckman, D. Szafir, and G.\ Sibley are with the Department of Computer Science, University of Colorado at Boulder. S.\ Julier is with the Department of Computer Science, University College London. }
\thanks{$^{\ast}$ Corresponding authors.}
\thanks{E-mail: {\tt\scriptsize[stephen.mcguire,nisar.ahmed]@colorado.edu}}%
}

\begin{document}
\maketitle
\thispagestyle{empty}
\pagestyle{empty}

%%%%%%%%%%%%%%%%%%%%%%%%%%%%%%%%%%%%%%%%%%%%%%%%%%%%%%%%%%%%%%%%%%%%%%%%%%%%%%%%
\begin{abstract}
Determining extrinsic calibration parameters is a necessity in any robotic system composed of actuators and cameras. Once a system is outside the lab environment, parameters must be determined without relying on outside artifacts such as calibration targets. %Since direct measurement to an image plane is impractical, 
We propose a method that relies on structured motion of an observed arm to recover extrinsic calibration parameters. Our method combines known arm kinematics with observations of conics in the image plane to calculate maximum-likelihood estimates for calibration extrinsics. This method is validated in simulation and tested against a real-world model, yielding results consistent with ruler-based estimates. Our method shows promise for estimating the pose of a camera relative to an articulated arm's end effector without requiring tedious measurements or external artifacts.

\end{abstract}

%%%%%%%%%%%%%%%%%%%%%%%%%%%%%%%%%%%%%%%%%%%%%%%%%%%%%%%%%%%%%%%%%%%%%%%%%%%%%%%%
\section{Introduction}
In robotics, data fusion between multiple sensors is frequently required in order to accomplish some task in an operating environment. Often, the relative locations between various components must be known to a high degree of precision; any error in relative pose propagates throughout the system without possibility for correction. These locations can be determined via idealized means, such as computer-aided design, or estimated means, such as motion capture systems or tag trackers. For many sensor types, the precise point of measurement may be buried within a housing or even within an integrated circuit, making direct measurement impractical. Furthermore, such idealized measurements may fail to capture assembly variations or installation error.

In this work we consider the goal of calibrating an assembled group of kinematically linked sensors and actuators as shown in \fig{over_sketch}. To support the idea of ``bolt together and go'' robotics, calibration should be determined by operating the assembly in its environment and analyzing the associated output without using calibration targets or precision artifacts. Such scenarios arise frequently in the operation of robotic arms, boom cameras, and other articulated sensor mounts used for capturing scientific data. A significant barrier in such operations is measuring the exact relative transforms between various sections of the kinematic chain, such as the transform from an imaging sensor to an articulated end effector. We develop an approach to calibration that does not rely on highly precise measurements for difficult-to-determine quantities; instead, we use a camera to observe the arm's motion and estimate the required quantities.

Our approach estimates relative poses between an actuator and camera by observing features on the actuator under structured motion. These features trace out circles in 3-D world space; we estimate a 3-D model to be projected into the image plane in order to find relative poses between an actuator and an observing camera. The present work is the first application of the tracking of features to model 3-D circles in space, rather than modeling elliptical paths in projective space.  This approach has many advantages related to noise robustness as described in Section \ref{sec:procedure}, and combines significant achievements in structure from motion, shape estimation, camera pose estimation, and self-calibration.

%Figure 1 notes:
% Make it a use case (little robot with an arm and a camera, camera with an over-the-shoulder view
% The paper is about cameras and arms
%
% Split into two figures: one for the specific problem to be solved 
%3D Rendering of a husky with an arm and a camera on it
%Include view from the camera showing the arm

%
\scaledfig{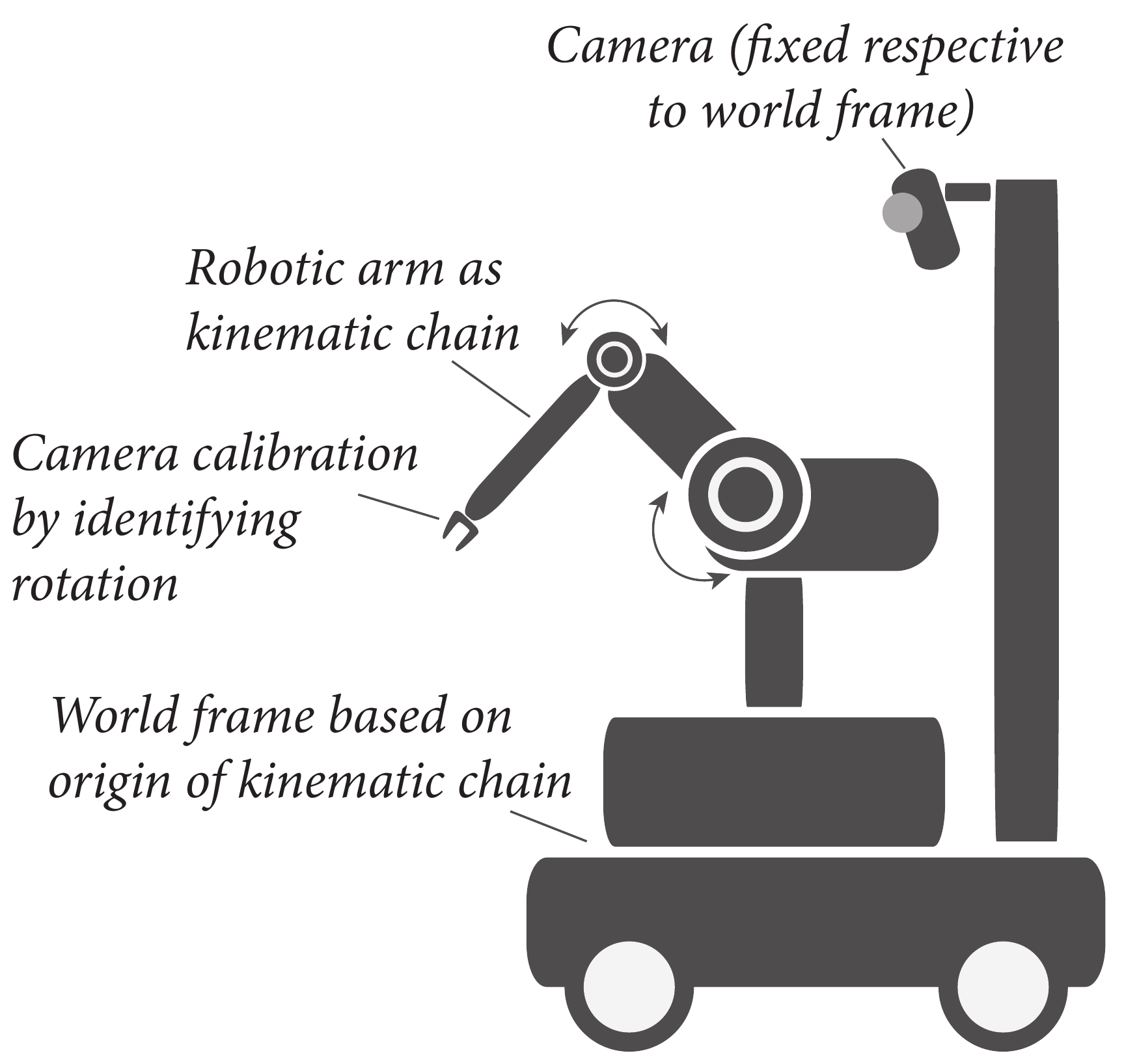}{over_sketch}{Overview of the hand-eye calibration problem on a mobile robot base. A fixed camera rigidly mounted to the base of a robot arm observes the arm's end effector movements.}{.45}

\scaledfig{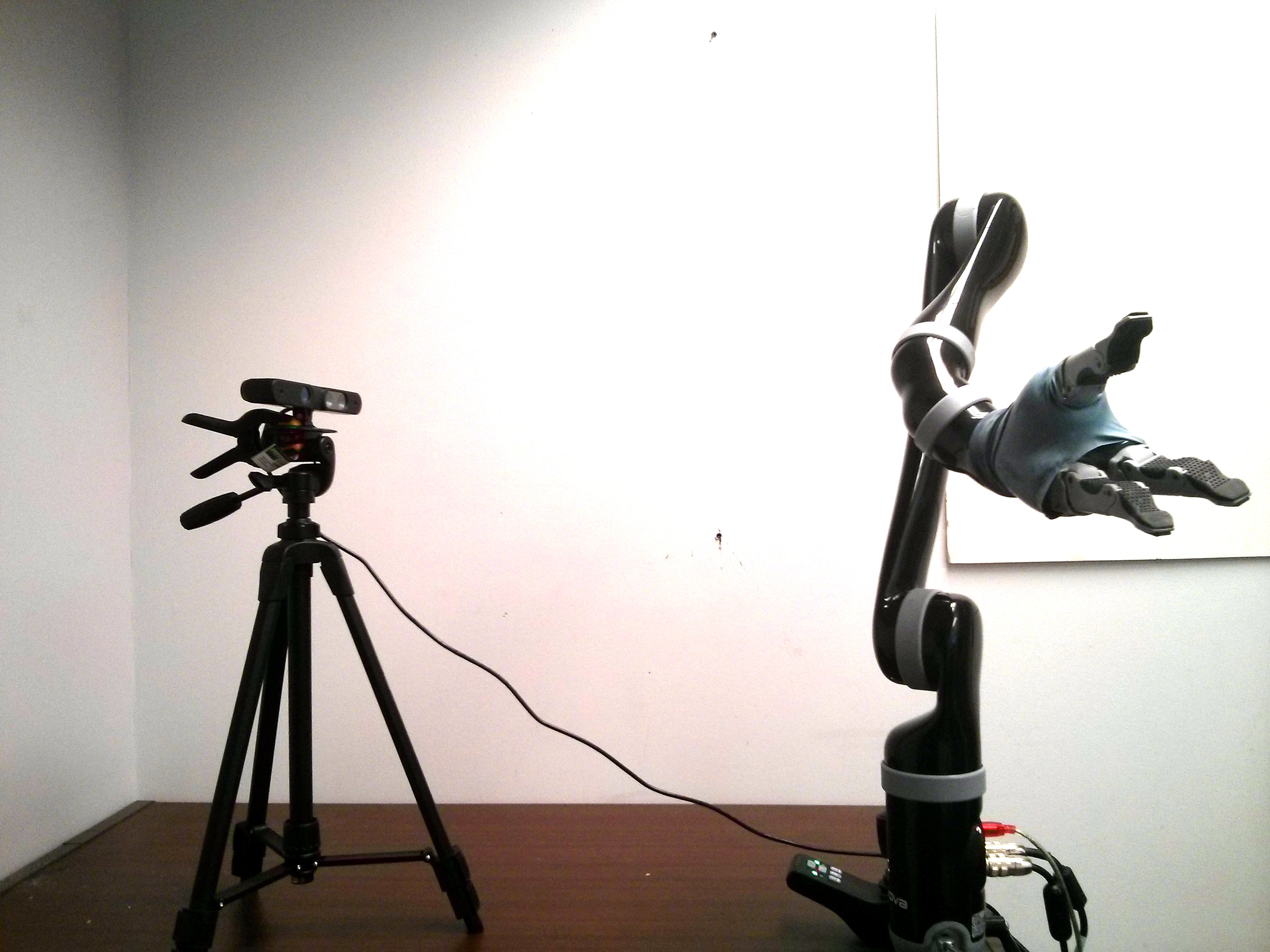}{exp_setup}{Real-world setup of the calibration problem showing a fixed camera rigidly mounted with respect to a robotic arm. In this experiment, a six-jointed Kinova Jaco2 arm is used.}{.45}

\scaledfig{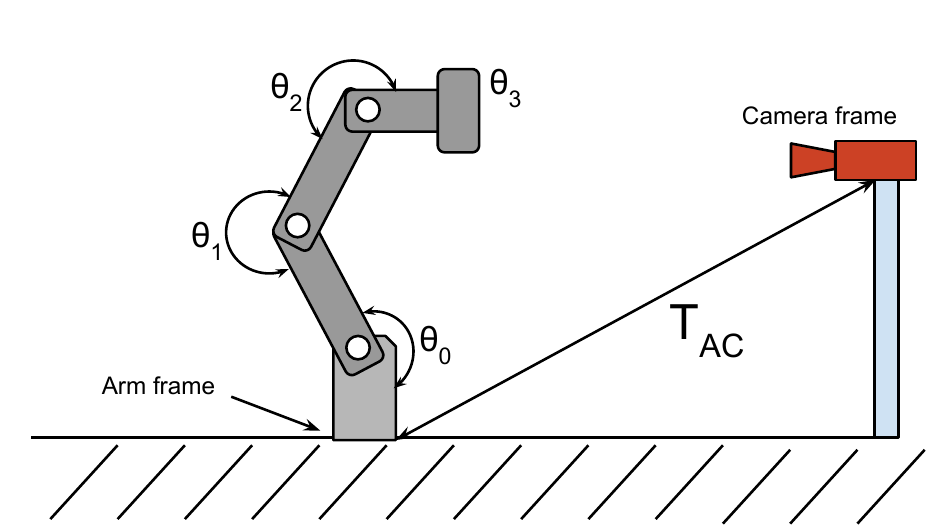}{prob_sketch}{Simplified sketch of setup. In this setup, the world frame is defined as the origin of the arm's kinematic chain. The camera is fixed with respect to the arm's kinematic origin via rigid transform $T_{AC}$. Note that our method does not assume coplanar rotations as suggested by the diagram.}{.45}

%Move to procedure (showing real-world)

The specific problem considered here is to determine extrinsic calibration parameters for a camera-arm system, such that the camera has a view of the end-effector of the arm, as shown in \fig{exp_setup}, with an idealized sketch in \fig{prob_sketch}. An arm has $i$ revolute joints yielding one degree of freedom per joint. Each arm joint produces a measurement $\theta_i$, representing the angular deflection of the revolute joint $i$. The arm has a well-known kinematic model, assumed to be precision manufactured. For this work, the final joint in the arm is a wrist-type joint with a rigidly-mounted end effector. The camera is assumed to be calibrated for intrinsic parameters via other means (e.g.\ \cite{Keivan2015}) and rigidly mounted via a fixed, but unknown, transform $T_{AC}$ to the origin of the arm's kinematic chain.

\section{Related work}
Camera-arm calibration is typically referred to as the ``hand-eye' problem in robotics and may be specified with either a stationary camera observing the motion of an arm or a mobile camera rigidly mounted to the end effector of an arm. 
An example of the current state of the art is from Pradeep et al.\ \cite{pradeep2014calibrating}, where precision calibration targets are moved via well-known robot kinematic models to recover relative transforms between cameras and the known kinematic models. Our work develops a technique for recovering these same transforms without the need for precision calibration targets by exploiting how a camera image changes under robot motion. Specifically, we are interested in wrist-like rotary motions. Through this type of arm motion, we can factor the hand-eye problem into several component problems: feature detection, feature tracking, geometry reconstruction, and finally pose reconstruction. Our work focuses on the geometry reconstruction and the pose reconstruction aspects of the factored hand-eye problem.

In our formulation of rotational geometry reconstruction, we are interested in recovering the parameters of a 3D circle from a set of observed image points. This is possible by leveraging an observation from Jiang et al.\ \cite{jiang2003geometry}, who originally established a method for reconstruction of a moving object with respect to a fixed camera under the assumption that the motion is constrained to rotation about a single axis. In that work, it was observed that individual points rotating about a single axis will trace out an ellipse under projection. The main focus of \cite{jiang2003geometry} relates to reconstruction from a single rotation; we use multiple independent rotations to establish the relative transform from the target object to the camera.

In \cite{sawhney1990description}, Sawhney et al.\ describe a method for reconstruction of objects under rotational motion, assuming that the camera undergoes the rotational motion attached to an arm observing a static scene. Sawhney develops much of the mathematical framework that will be used in our method. In \cite{marinello2008critical}, Fremont and Chellali address how to create a 3-D reconstruction of an object by rotating it about an axis, but does not address how to estimate the pose of such an object with respect to an arbitrary image plane. In \cite{liu2014relative}, Liu and Hu use a fixed camera to observe a cylindrical spacecraft and estimate pose using a known CAD model to match ellipses to metric features. Their work relies on static imagery and resolves scale using the CAD model; in our formulation, we have no information about the configuration of the end effector.

Hutter and Brewer \cite{hutter2009matching} used the approach of approximating elliptical shapes in the image plane corresponding to true circles in 3-space \cite{sawhney1990description, kanatani19933d,chen2004camera} to estimate the pose of vehicle wheels from segmented images for self-driving car applications. Their work recovers the wheel's rotation axis in order to estimate steering angle. Our work accomplishes a parallel goal of recovering the end effector's rotation axis in order to estimate properties of the larger system. Another application is in \cite{alismail2012automatic}, where pose estimation from an image of a well-known ellipse is used to calibrate a laser-camera system. However, the parameters of the underlying circle must be known a priori in the form of a known calibration target.

Lundberg et al. \cite{lundberg2014intrinsic} pose a similar problem setup, but use a single well-known feature on the end effector. This well-known feature is precisely positioned in each of a series of frames to form a `virtual' calibration target. A calibration target is thus assembled in a point-wise manner, allowing calibration to continue using conventional techniques. A key difference between their solution and ours is the uniqueness and the prior knowledge of the well-known feature; our solution can utilize ambient features discovered at runtime.

Also connected is that of Forsyth et al.\ \cite{forsyth1991invariant}, where camera pose was estimated by observing images of well-known conics. In contrast, we observe conics by tracking features of a rotating object over time; a critical distinction is that the world geometry of the observed conics is not well-known. Borghese et al.\ \cite{frosio2012linear} also applied Forsyth's work for pose estimation, but only applied this to rotated calibration targets and not to generated conic paths in 3-D space resulting from object rotation.

\section{Procedure}
\label{sec:procedure}
\begin{comment}
\section{Nomenclature}
Vector quantities are in bold lower-case, such as \textbf{x}. 
Matrix quantities are in bold upper-case, such as \textbf{T}.
Scalars are in lower-case italics, such as \textit{x}.
Sets are in capital italics, such as \textit{S}.

Chirality is right-handed, Euler angles are zxy. 
\end{comment}

The goal of our method is to estimate the camera's location relative to the base of the arm's kinematic chain. First, images are captured of the end effector rotating about the axis of the last joint in the arm while the remainder of the arm is motionless. While the arm is rotating, feature points (described in Sec. \ref{sec:arc_ident}) on the end effector are tracked and associated between frames to produce a set of arcs in the image plane. During an initial optimization step, the observed axis of rotation is estimated by requiring that each arc's center of rotation be collinear, with independent radii and distance along the rotation axis from an arbitrary starting point. Meanwhile, the arm's joint angles $\theta_i$ are captured to calculate the arm's expected axis of rotation in the world frame. This procedure is repeated a number of times $n$, where the arm is repositioned to a different pose and the process is repeated. After all measurements are collected, the position of the camera in the arm frame is estimated. This technique is presented in video form in the Supplementary Materials.

We use the encoding of a pose $\bm{p}\in \mathbb{R}^6$ as composed of six elements: $\rowvec{x,y,z,\phi,\theta,\psi}$ (the latter three quantities representing roll, pitch, and yaw respectively). Each measurement observes the $x,y$ position in the image plane of $m$ feature points, as well as the arm angular measurements $\theta_i$. The set of all unique observed feature points is $\textit{L}$. A $4 \times 4$ homogeneous transform giving the location of object $A$ in coordinate frame $B$ is denoted $\bm{T}_{BA}$. %With this convention, chains of transformations may be premultiplied and frames canceled in a transitive manner, such as $\bm{T}_{CA} = \bm{T}_{CB}\bm{T}_{BA}$.
Estimates of a particular quantity $P$ are denoted by $\hat{P}$. Parameters to be estimated comprise the pose $\bm{p}$ corresponding to the location of the camera in the arm frame, represented alternatively as $\bm{p}_{AC} \in \mathbb{R}^6 $ or $\bm{T}_{AC}$ as a $4 \times 4$ homogeneous transformation. These two representations will be used interchangeably via the Lie SE(3) $\exp()$ and $\log()$ operators where needed. 

%Fig 2 notes:
% Message: Circles get viewed as ellipses
% Split into two: camera view and overview with variables of geometry of the scene
% Pictorial Problem definition
% More detail, formalize the problem
%

%\scaledfig{figures/cam_proj_ellipse}{cam_proj_ellipse}{Diagram of projective geometry of an ellipse. An ideal circle is observed as an ellipse.}{.3}

The estimation procedure follows several steps: 1) arc identification, 2) circle estimation, 3) estimation of the rotation axis by vision, 4) rotation axis estimation using forward kinematics, and 5) estimation of $\hat{T}_{AC}$.

% \pseudo{gen_steps}{
% \algblock[Name]{Start}{End}
% \Start
% \State Capture images
% \State Extract arcs
% \State Fit 3D circles in arbitrary space
% \State Generate test points
% \State Minimize residual
% \End
% }{General calibration procedure}

\subsection{Arc Identification}
\label{sec:arc_ident}
To establish arcs, ambient features on the end effector are tracked across frames. The choice of feature descriptor is arbitrary, so long as the end effector can be tracked across more than three frames to be able to fit a projection of a 3-D circle to the track. The reference implementation utilizes OpenCV's simple blob detector as input to the Lucas-Kanade optical flow algorithm to obviate the effect of local minima. Tracks that are shorter than a minimum threshold are discarded, as are tracks that do not encode sufficient motion between frames. As features potentially rotate out of view, new features are detected as potential tracks. The use of ambient features forms the novelty of our work, eliminating the need for precision targets and/or well-known features. 

\subsection{Rotation axis identification (vision)}
Once tracks are assembled, an initial optimization step is performed to recover the shared rotation axis of the generated ellipses. The choice of coordinate axes corresponds to the image frame, with $x$ right, $y$ down, and $z$ forward. In this optimization step, the parameter vector $\mathbf{l}_k$ consists of 5+2\textit{m} terms: an $(x,y,z)$ point on a 3-D line, angles $\theta$ and $\phi$ indicating the rotation along XY and YZ axes, and a radius from the line and displacement from the origin point for each ellipse to be fit. This parameterization enforces a coaxial constraint such that the centers of every circle in 3-D are collinear. In this stage, the camera is set at $(0,0,-1,0,0,0)$ with respect to the origin. Initial conditions for the 3-D circle optimization are set to $(0,0,0,0,0)$, with each potential arc having incremental displacement and unit radius. The coordinates to be recovered are in an arbitrarily transformed up to a scale space. 

Candidate points on circles corresponding to the parameter vector are then generated and projected into the image plane according to the given camera model's projection matrix. For each detected feature's arc, an ellipse is then fit to these projected points belonging to the parameter vector's 3D circle using a direct least squares method \cite{fitzgibbon99}. A residual is then composed by summing the distance between projected points and the candidate ellipse in projective space for each feature track. The use of a coaxial constraint on the underlying model used to generate fit ellipses aids in noise robustness, as a poor track can corrupt an individual ellipse fit.

For each measurement $k$, the candidate parameters $\mathbf{l}_k$ identify a line in 3-D; points are generated along this line, projected into the image plane, and a best-fit line in 2-D is recovered; this line represents the projection of the axis of rotation into the image plane.

These lines are stored in point-slope form, yielding an observation $z_k = (m_k,b_k)$.

\subsection{Rotation axis identification (arm)}
Given that the kinematic chain of the arm is provided, without loss of generality we consider a single joint of the arm. With a revolute joint, the transition through the joint is characterized by seven parameters: six to identify the fixed mechanism of the arm leading into the joint (denoted link $L$), and one to adjust the output face based on actuator angle (denoted $T_{\theta_i}$). Each transform is represented in the parent joint's coordinate system; for joint $i$, the kinematic chain is therefore represented in the coordinate system of joint $i-1$ as:

\neweq{arm_kinematics}{
T_{(i-1)(i)} = L_{i}T_{\theta_i}
}

Note that $T_{\theta_i}$ only encodes a single rotation; the axis of rotation of joint $i$. Therefore, any points of the form $\colvec{0\  0\ z'}^T$ will lie on the rotation axis of joint $i$, in joint $i$'s coordinate frame. World coordinates of a set of such points may be obtained by applying the arm's kinematic chain using homogeneous transformations.

\subsection{Camera to arm rigid transform estimation}
The estimated rigid transformation between the camera and the base of the arm $\hat{\bm{T}}_{AC}$ is estimated by applying bundle adjustment over all measurements $n$. 

Modeling the image noise as zero-mean Gaussian results in a maximum-likelihood estimator, which can be stated as a nonlinear least squares optimization problem of the form
\neweq{nonlinear_ls}{
\bm{r = e(z,x)}R^{-1}\bm{e(z,x))}^T
}
where $\bm{e(z,x)}$ is function of observations $\mathbf{z}$ and model parameters $\mathbf{x}$ that produces a residual error vector, and $R$ is a block-matrix of weights in the projected axis from the vision system corresponding to the uncertainties in the axis recovery.  For this problem, the parameters $\mathbf{x}$ to be recovered form a vector $\rowvec{x, y, z, \phi, \theta, \psi}$
representing the $\mathbb{R}^6$ form of $\hat{\bm{T}}_{AC}$, while the observations $\mathbf{z}$ are the stacked coefficients of the line fits $z_k = (m_k, b_k), k=1...n$ describing the axis of rotation for each of $n$ measurements. Since the recovered projected centers corresponding to each feature track are each located some unknown distance from the origin of the joint immediately prior to the end effector, we cannot directly compare known world geometry of the end effector to the observed axes of rotation. However, we exploit the fact that the true axis of rotation must be collinear with the observed axis of rotation to form a residual function $\bm{e}$.  

For each of $n$ measurements $k$, several test points of the form $\colvec{0\  0\ z'}^T$ are generated in the last joint's coordinate frame. A minimum of two test points are required to define the line representing the projection of the arm's axis of rotation in the image plane. While more points can be used, the assumption of a rectified camera model guarantees that any additional points will be collinear. These $j$ test points are first transformed into world coordinates using the arm's kinematic model and then projected into the image plane using the current parameter vector $\bm{x}$. For each measurement $k$, the previously identified line parameters $\colvec{m_k\ b_k}$, the distance $d_{k,j}$ between each projected test point $t_j = \colvec{u_j\ v_j}$ and its associated projected axis of rotation $a_k$ is calculated by:
\neweq{distance_eq}{
d_{k,j} = \frac{-m_ku_j+v_j-b_k}{\sqrt{m_k^2+1}}
} as given in \cite{line_distance}.

The distance metric of \Eq{distance_eq} is used to assemble the residual vector $\bm{r}$ by concatenating over all test points and all measurements:
\neweq{res_vector}{
\bm{r} = \rowvec{d_{11}\ d_{12}\ ...\ d_{1j}\ d_{21}\ d_{22}\ ...\ d_{2j}\ ...\ d_{n1}\ d_{n2}\ ...\ d_{nj}}
} We implement our optimization using Levenberg-Marquardt with random restarts to avoid local minima.

To avoid numerical instability resulting from the use of finite differences, we apply the technique of \cite{martins2003complex} to determine the value of the Jacobian at the estimate $\mathbf{\hat{x}}$, defined as 
\neweq{pose_jacobian}{
	\mathbf{J} = \frac{\partial{\mathbf{r}}}{\partial{\mathbf{x}}}\bigg|_{\hat{\mathbf{x}}}
} Once $\mathbf{J}$ is determined, the diagonal terms of $(\bm{J}^T\bm{J})^{-1}$ estimate the variance in each recovered component of $\hat{\bm{T}}_{AC}$.

\subsection{Measures}
We evaluate our simulation results through two sets of measures. The first set evaluates the quality of the rotation axis identification from feature track points, while the second evaluates the quality of the camera pose reconstruction. Rotation axis identification quality is measured by relative error in the 2D line parameters $\rowvec{m_k, b_k}$ for each measurement $k$; these errors are expressed in $\text{pixels}^{-1}$ and pixels respectively. The camera pose reconstruction quality is measured by the relative error in 3D position parameters $\rowvec{x,y,z,\phi,\theta,\psi}$, expressed in units of meters and radians.

\section{Results}

\subsection{Simulation}
In simulation, observed arc data are generated against a given camera location and arm end effector positions. An example of a simulated camera view is \Figure{sim_ellipse_proj_view}, showing the progression from source data to the final measurement that will be used in the optimization, a line through the projected circle centers.

% make x-y spacing equal
% zoom in a bit
% font size
%
%

\scaledfig{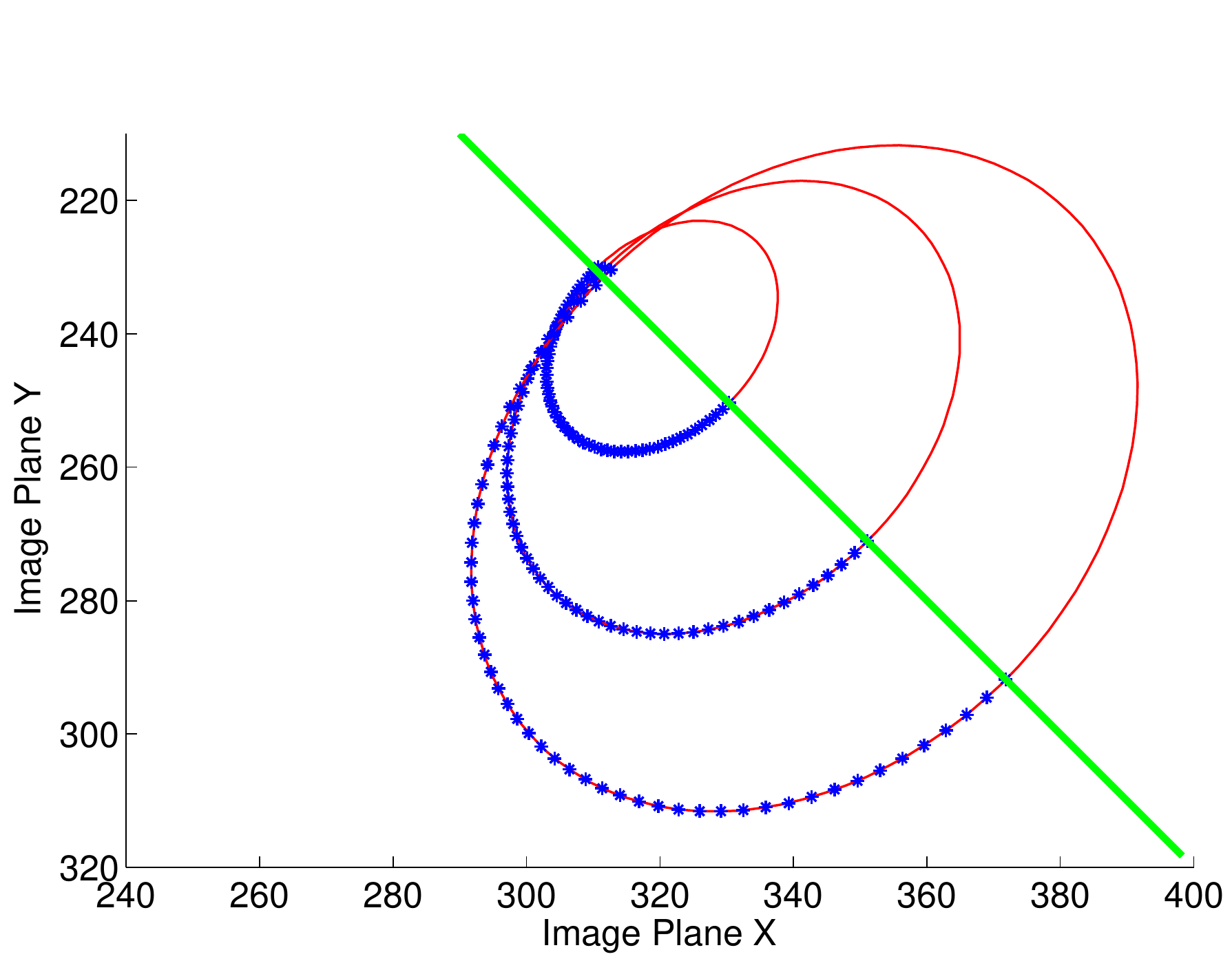}{sim_ellipse_proj_view}{Simulated projected view of a set of three feature tracks. Source track data are in blue stars, fitted ellipses in solid red, and projected axis of rotation through circle centers in green.}{.48}

To characterize how tracking error propagates through the estimation technique, a Monte Carlo simulation was run at four different levels. Since two stages of optimization are present, we present results at both the ellipse fit stage and the final camera position fit stage. Zero-mean Gaussian noise at $\sigma^2 = \{0.1,0.5,1.0,1.5\}$ px$^2$ was added to the projection of simulated ellipse points into the camera image plane. Six rotation observations were simulated; the error distributions are shown in \fig{linefit-pair}.
%\leftfig{figures/linefit-0-1}{linefit-0.1}{Error distribution of rotation axis fit with $\sigma^2=0.1$ px$^2$ noise}
%\leftfig{figures/linefit-0-5}{linefit-0.5}{Error distribution of rotation axis fit with $\sigma^2=0.5$ px noise}
%\leftfig{figures/linefit-1-0}{linefit-1.0}{Error distribution of rotation axis fit with $\sigma^2=1.0$ px noise}
\scaledfig{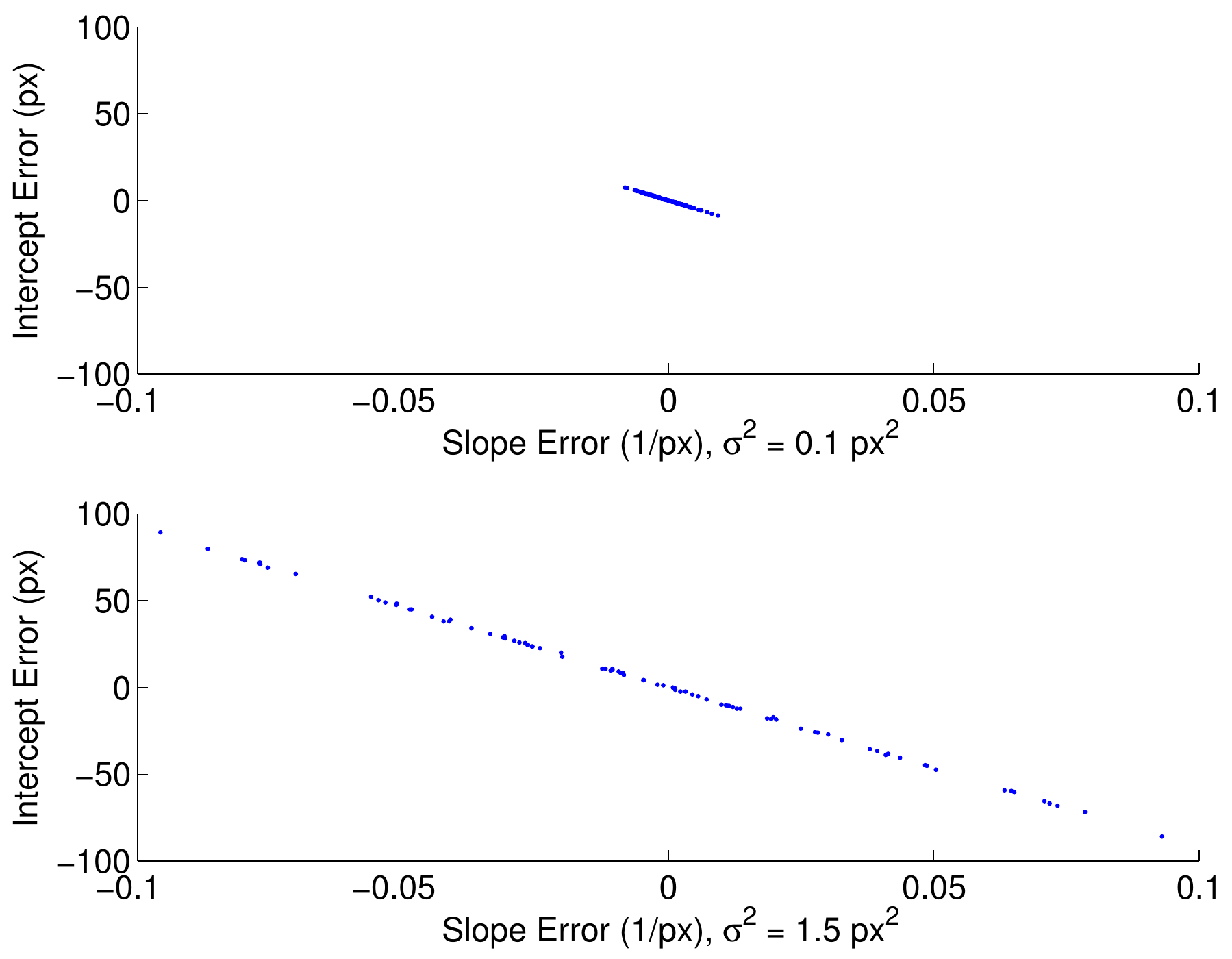}{linefit-pair}{Error distribution of rotation axis fits in with $\sigma^2=0.1$ px$^2$ and $\sigma^2=1.5$ px$^2$ noise showing error correlation and impact of increased tracking noise}{.49}

Each of these simulation runs was then pushed through the second optimization step to recover the 3D pose of the camera. The error distributions in the Cartesian directions are shown in \fig{cart-0.1} and \fig{cart-1.5}, while the angular errors are shown in \fig{ang-0.1} and \fig{ang-1.5}.

\scaledfig{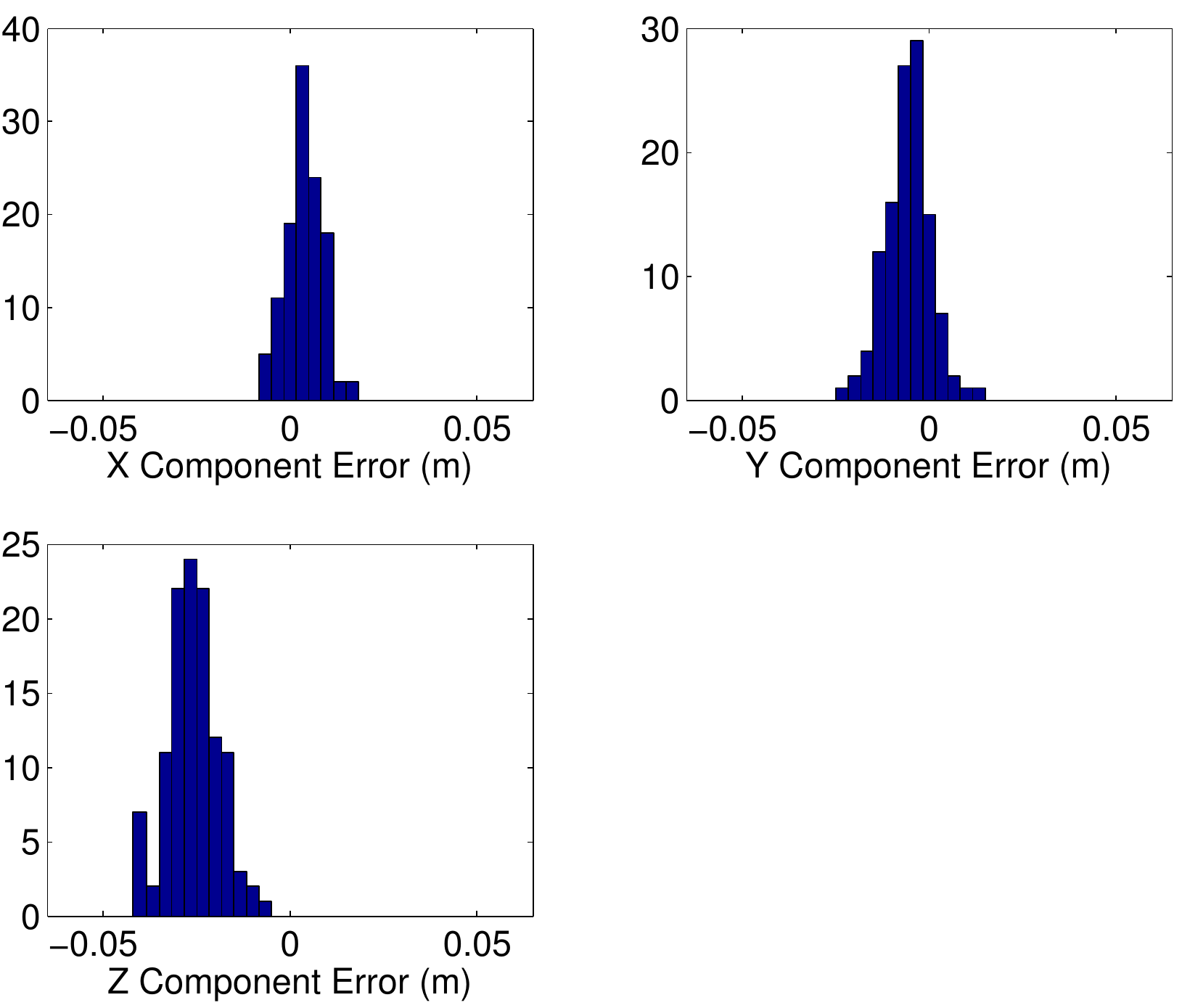}{cart-0.1}{Cartesian position error distribution with $\sigma^2=0.1$ px$^2$ noise}{.49}
%\leftfig{figures/cart-0-5}{cart-0.5}{Cartesian position error distribution with $\sigma^2=0.5$ px noise}
%\leftfig{figures/cart-1-0}{cart-1.0}{Cartesian position error distribution with $\sigma^2=1.0$ px noise}
\leftfig{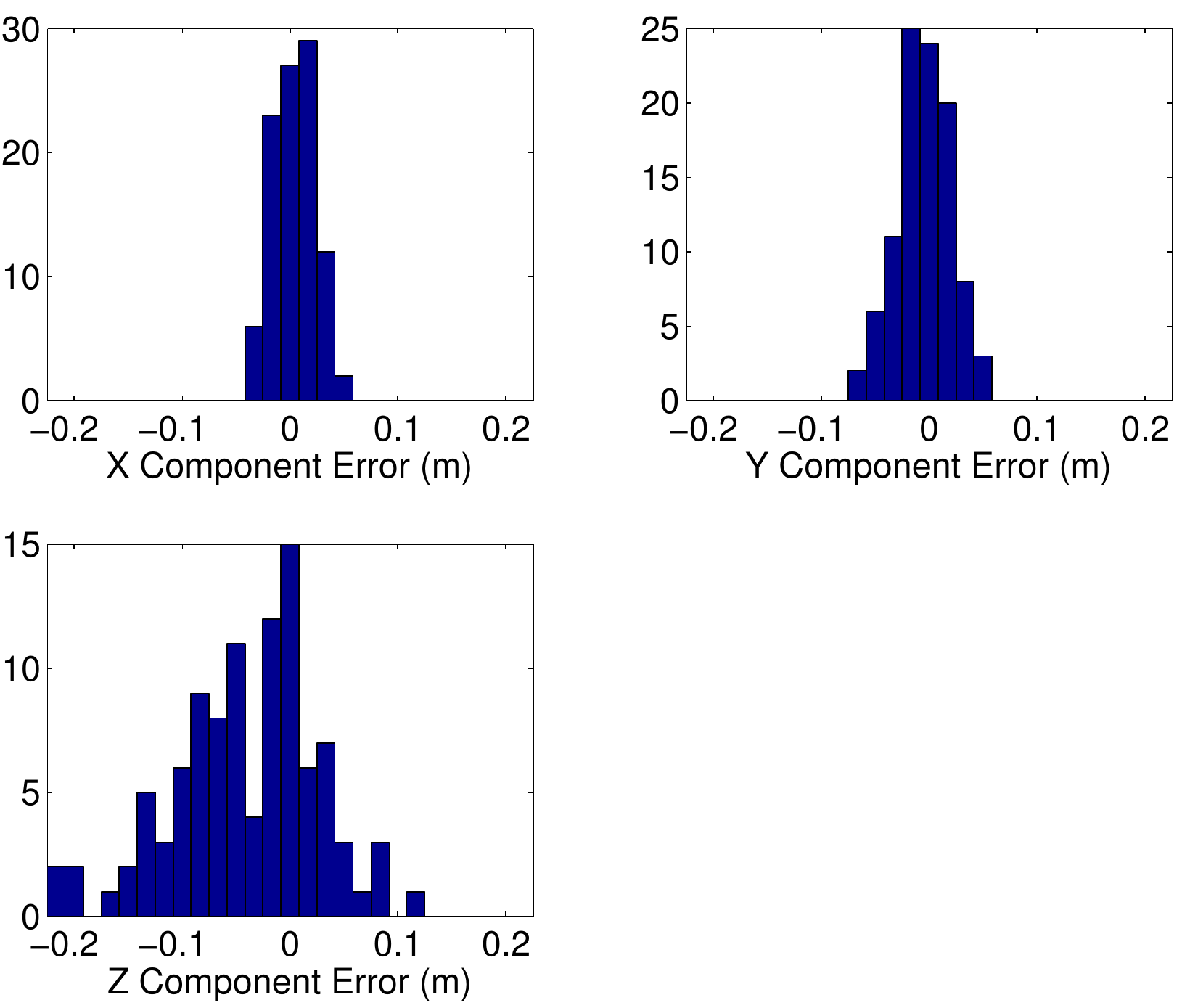}{cart-1.5}{Cartesian position error distribution with $\sigma^2=1.5$ px$^2$ noise}

\leftfig{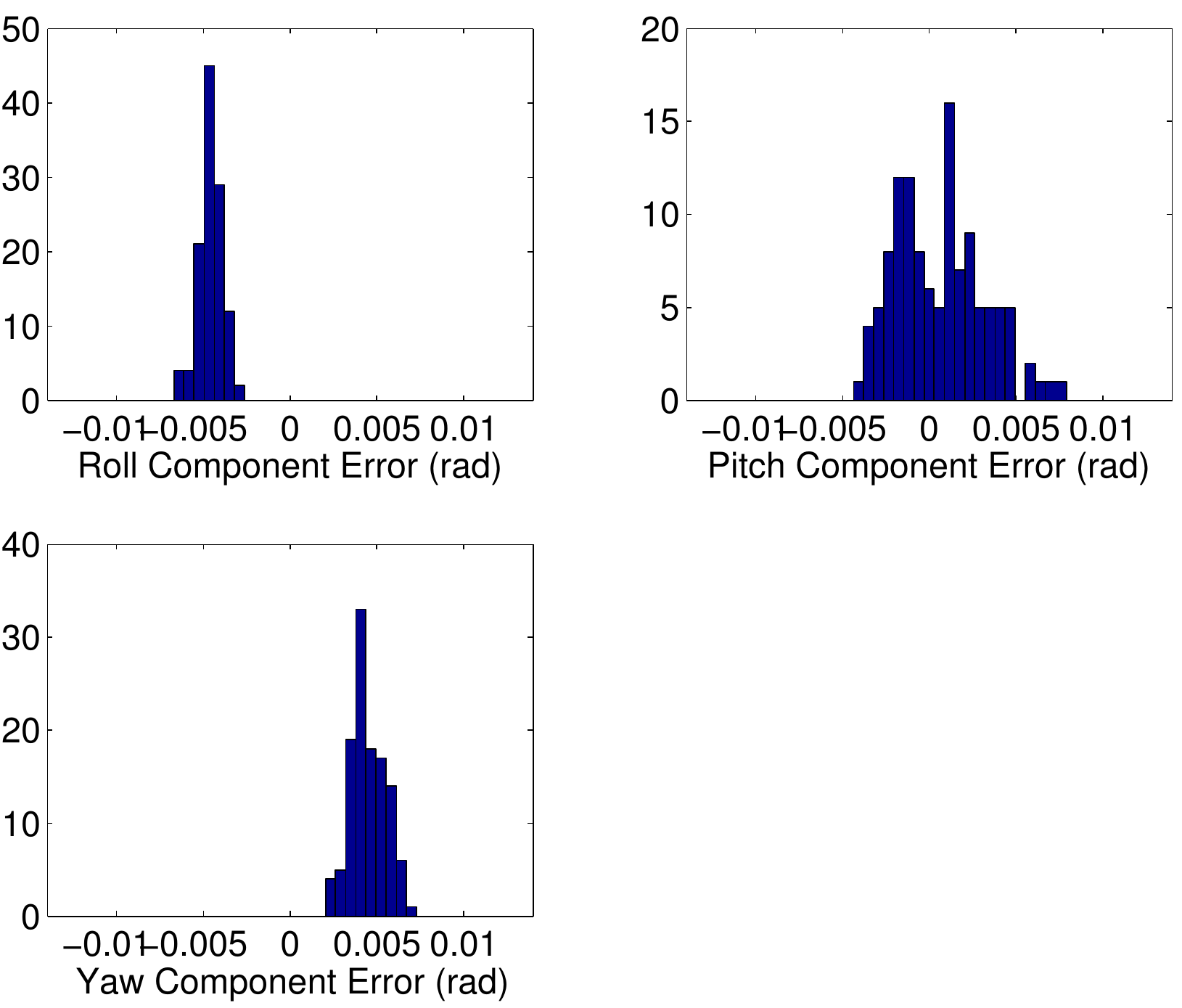}{ang-0.1}{Angular error distribution with $\sigma^2=0.1$ px$^2$ noise}
%\leftfig{figures/ang-0-5}{ang-0.5}{Angular error distribution with $\sigma^2=0.5$ px noise}
%\leftfig{figures/ang-1-0}{ang-1.0}{Angular error distribution with $\sigma^2=1.0$ px noise}
\leftfig{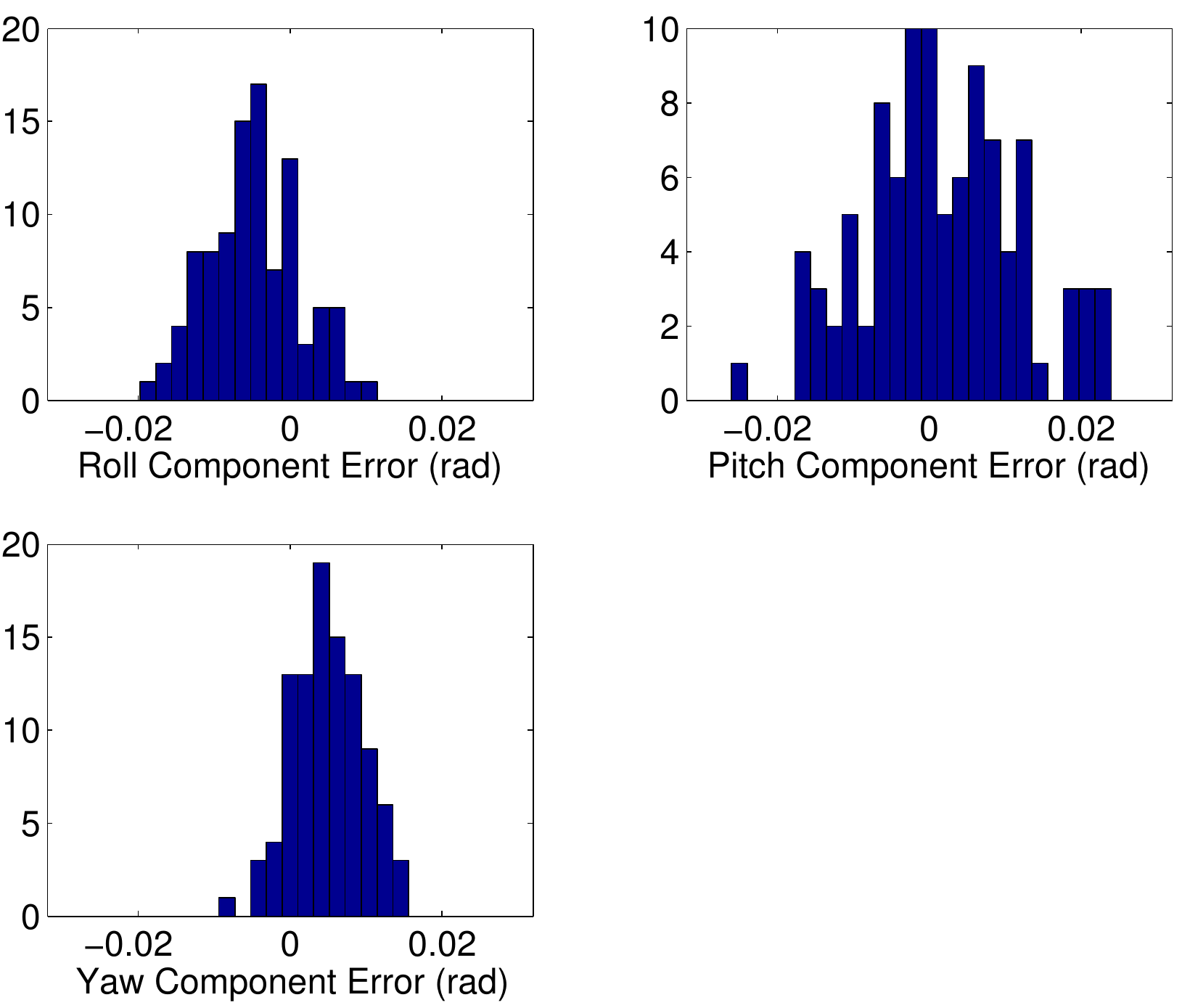}{ang-1.5}{Angular error distribution with $\sigma^2=1.5$ px$^2$ noise}

\subsection{Physical System}
The system was tested using an Asus Xtion Pro RGB-D camera and a Kinova Jaco2 arm, shown in \Figure{exp_setup}. Calibration of camera intrinsics occurred offline, while the Kinova Jaco2 forward kinematics were derived from the manufacturer-provided model. The arm's end effector was equipped with a feature-rich covering to ensure adequate ambient features are available to be tracked, shown in \fig{quad_view}. As features were tracked in the image, track data was assembled. In the real system, track data is not as smooth as the simulation; additional processing was implemented to detect jumps in the track that were inconsistent with a smooth arc. A minimum track length was established to remove poor tracks. Measurements were collected independently, with the estimation running offline. Each measurement consisted of approximately sixty seconds of rotation in the wrist joint, followed by an arm reposition for the next measurement. 

\scaledfig{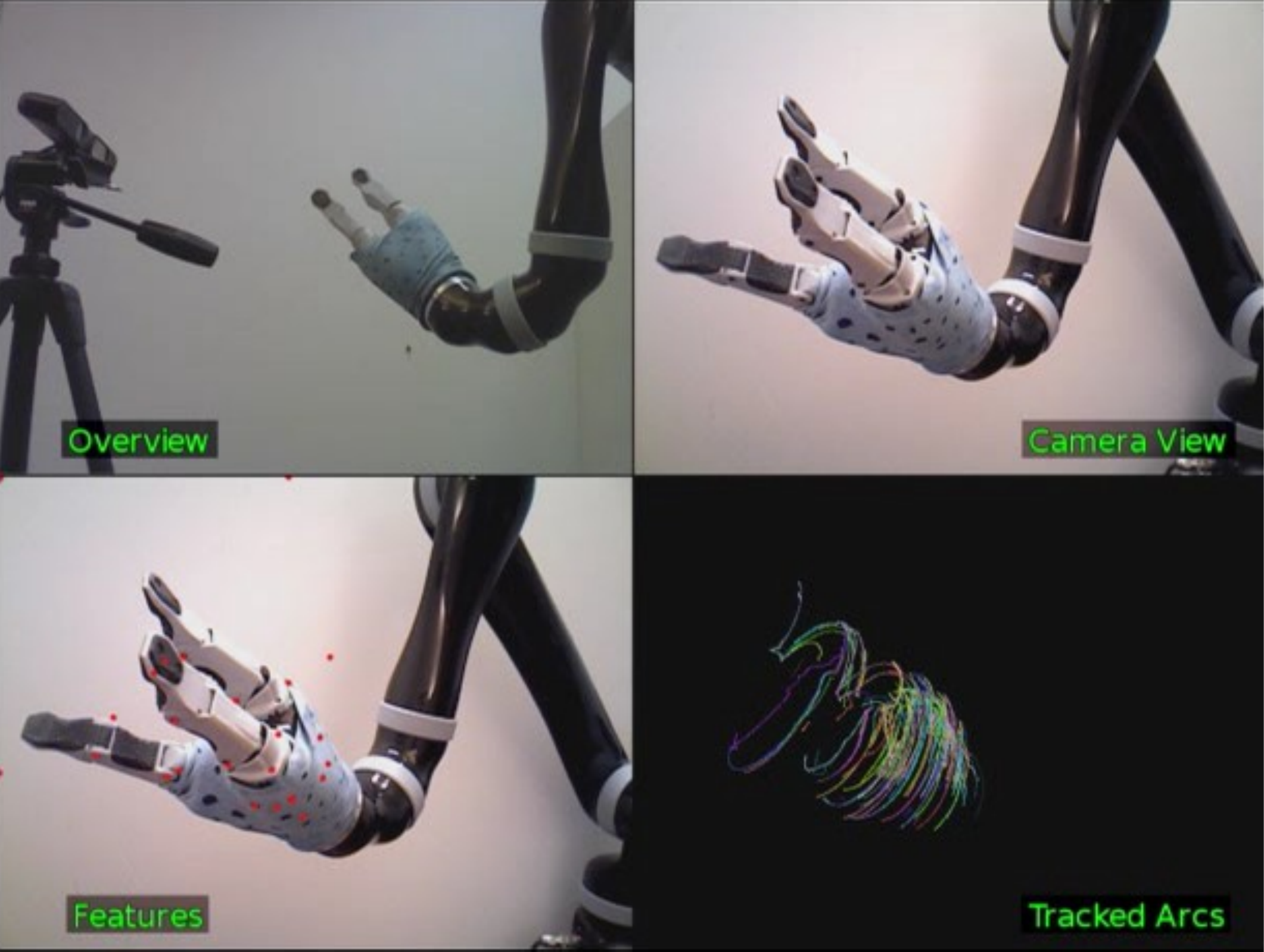}{quad_view}{Four views of the real system under test, showing an overview, the camera view, the detected features, and the tracked arcs}{.47}

%\scaledfig{figures/arc_tracks_wht}{track_example}{Example tracks measured from the real system using a 640x480 pixel RGB camera. Each feature track is drawn in a different color. Features on different portions of the end effector appear as ellipses of varying sizes.}{.4}

%Potentially cut - what does it add? 
%\scaledfig{figures/real_sys}{real_results}{A final optimization produced from the real system using four measurements. Rotation axes estimated from vision are in solid red, while points along the arm's axis of rotation are in blue stars.}{.4}

After optimization completed, we were able to recover a pose consistent with ruler-based estimates of $\hat{\bm{T}}_{AC}$. Our covariance estimate reported errors on the order of 0.1mm; these results appear to be excessively optimistic, given the relatively low quality of arc tracks used as observations. Reasons for this optimism could include the omission of error weights in the estimation of $(\bm{J}^T\bm{J})^{-1}$, implying that all measurements are equally trustworthy.

\section{Discussion}

In simulation, our technique was able to recover reasonably accurate estimates of camera position even in the presence of tracking noise. Given the minimal time required to gather a calibration dataset and modest algorithm execution time (approx 5 minutes with a Matlab implementation), our technique seems practical to implement in real-world scenarios. We were able to demonstrate the function of our system aboard a representative real system with varying success. 

In the line fit distributions, particularly at higher noise levels such as \fig{linefit-pair}, we observe that the error terms are strongly correlated; we hypothesize that this correlation is due to the external constraint that all ellipses be coaxial.  The noise characteristics appear to be dependent on the viewing angle of the camera with respect to the rotation axis of the arm. When recovering the line fit parameters for the measurement data associated with the wider distributions in \fig{linefit-pair}, the optimizer terminated with a summed residual value several orders of magnitude higher than the summed residuals of the narrow distributions. By treating large residuals as outliers, the overall robustness of the estimation technique could be improved.

When the line fit errors are propagated through to the camera pose recovery step, we note that several position estimates appear to be biased, for example, the Z component error in \fig{cart-0.1}. These error propagation results lead us to conclude that an uncorrelated and unbiased Gaussian error model in output noise is not appropriate for this estimation technique. The variance of the expected errors are within several centimeters even at the higher noise level. Based on these simulation results, we believe that this technique is worth further investigation.

While our method has shown promising results, several important areas remain for further investigation. The effect of each of these areas on our results is demonstrated in the Supplementary Materials.

\subsection{Measurement constraints}
Several conditions on arm pose have been identified as necessary (but not sufficient) to have a convergent optimization. For example, in a single crossing scenario, the end effector of the arm was commanded to have the same $\colvec{x\ y\ z}$  position with different orientation values. This setup represents a  degenerate pathological problem because the location of the camera is underdetermined. Possible locations for the camera lay on a line in 3D space perpendicular to the 2D projection of the crossing point; since the residual computation only considers distance from a test point to the associated rotation axis, all camera positions yield the same residual value. Camera orientation can be recovered, but position cannot. To avoid this condition, a minimum of three measurements are required, with one measurement having a different position from the other two. 

%\scaledfig{figures/fail_single_crossing}{fail_single_crossing}{Failure due to single point of crossing of rotation axes. Red lines are identified rotation axes from vision (one per measurement), while blue stars are test points along the arm's rotation axis.}{.48}

\subsection{Differences between simulation and real robot}
The performance under simulation appears to recover the camera position more accurately than the use of a real system; we have several hypotheses to explore these differences. In the real system, a physical camera was used with a basic linear camera model that fails to completely replicate the ideal camera used in simulation. There is also quantization error in the camera vs simulation, particularly in the generation of arc portions used in the ellipse fit stage. In simulation, these arc points are carried through with double-precision floating point values throughout; while subpixel resolution techniques in real imagery can help to mitigate this effect, further work is required to improve the fidelity of the simulation. We studied error in tracking by introducing zero-mean independent Gaussian noise into the arc track data image coordinates; observing the real tracks produced in our reference implementation, shown in \fig{quad_view}, a Gaussian noise model may not be appropriate.

\subsection{Optimization initial values}
In both stages of optimization, testing has revealed a significant dependency on initial conditions to the optimizer to yield a convergent solution due to degeneracies in the 3D circle fitting procedure; this phenomenon is very common in SLAM algorithms as noted by \cite{newman2005slam}. Since the residual is defined as the distance between the projected ideal circle and the observed arc track, there is no analysis of curvature to ensure that the projected ideal circle is approximately oriented at initial evaluation of the residual function.

\subsection{Known kinematics}
This approach relies on the manufacturer-provided model of the arm kinematics in order to resolve the rotation axis at the end effector with respect to the base of the arm as truth. This assumption could be improved by adding optimization parameters such as angular encoder bias to the various joints of the arm, to allow for installation errors. Other types of deviations from the manufacturer model, such as damage or wear, offer significant challenges in modeling but might be amenable to an online error analysis such as in \cite{Keivan2015}.

\subsection{Precalibrated camera}
For our optimization to be successful, we require known camera parameters and linearly rectified camera data as input. An important observation is that monocular self-calibration techniques such as \cite{KeivanS15} and our technique are not exclusive; as self-calibration relies on motion in the environment and the arm is rigidly mounted to the camera, the arm will appear as a static obstruction. This requires the arm to be held stationary during calibration, while the reference implementation of the arc tracker requires the background to be stationary.

\subsection{Tracker noise}
In the reference implementation, a basic tracker was used that did not attempt to enforce any motion dynamics on the detected tracks. An improvement on this work would include integrating a more advanced tracker that can incorporate a motion model to smooth out detected arcs and make the tracker more robust to noisy image data.

\subsection{Path planning for calibration poses}
An important unanswered question in our technique is the method of determining what constitutes a `good' pose for use in calibration. Ideally, a statistical metric could be determined to evaluate the information content of a new pose given the set of existing poses; this metric could then be used to develop a path planner that identifies a series of most valuable poses for reducing error in the final estimate of $\hat{\bm{T}}_{AC}$ under constraints of robot kinematics and camera field of view. 

\subsection{Coordinate system consistency}
While we have the convenience of defining our coordinate frames without regard to external needs, a real system must take into account a wide range of sensor data that may or may not even use the same coordinate or rotation convention. Multiple coordinate system standards pose a hazard to interpretation and development of calibration routines and reduce the generality of our technique.

\subsection{Incorporating measurement weights}
In future work, we envision using the residual value of the rotation axis identification stage as a means of weighting the relative value of that measurement in the pose reconstruction stage. The residuals of all measurements could be set on a scale to aid in outlier rejection, such that poor tracking in one particular measurement does not contaminate the overall result. 

\section{Conclusions}
We have presented a method for calibrating an articulated arm with a wrist joint to a camera without requiring calibration targets, instead relying on structured arm motion and ambient features on the end effector. Our method requires no knowledge of the mechanism of the end effector or unique features to be present. We have validated our results in simulation and demonstrated with real-world data yielding promising results. A Monte Carlo analysis of error propagation verifies that small errors in the quality of feature tracks do not cause the resulting position estimates to degrade dramatically. Further testing is required to explore system degeneracies and validate performance in realistic environments.

\section{Acknowledgments}
S.\ McGuire is supported by a NASA Space Technology Research Fellowship through grant number NNX15AQ14H. This work is also supported by the Toyota Motor Corporation.

\bibliographystyle{IEEEtran}
\bibliography{master.bib}
\end{document}